# PIIBench: A Unified Multi-Source Benchmark Corpus for Personally Identifiable Information Detection

**Pritesh Jha**

*Independent Researcher*

priteshjha2711@gmail.com

## Abstract

We present PIIBench, a unified benchmark corpus for Personally Identifiable Information (PII) detection in natural language text. Existing resources for PII detection are fragmented across domain-specific corpora with mutually incompatible annotation schemes, preventing systematic comparison of detection systems. We consolidate ten publicly available datasets spanning synthetic PII corpora, multilingual Named Entity Recognition (NER) benchmarks, and financial-domain annotated text, yielding a corpus of 2,369,883 annotated sequences and 3.35 million entity mentions across 48 canonical PII entity types. We develop a principled normalization pipeline that maps 80+ source-specific label variants to a standardized BIO tagging scheme, applies frequency-based suppression of near-absent entity types, and produces stratified 80/10/10 train/validation/test splits preserving source distribution. To establish baseline difficulty, we evaluate eight published systems spanning rule-based engines (Microsoft Presidio), general-purpose NER models (spaCy, BERT-base NER, XLM-RoBERTa NER, SpanMarker mBERT, SpanMarker BERT), a PII-specific model (Piiranha DeBERTa), and a financial NER specialist (XtremeDistil FiNER). All systems achieve span-level F1 below 0.14, with the best system (Presidio, F1=0.1385) still producing zero recall on most entity types. These results directly quantify the domain-silo problem and demonstrate that PIIBench presents a substantially harder and more comprehensive evaluation challenge than any existing single-source PII dataset. The dataset construction pipeline and benchmark evaluation code are publicly available at https://github.com/pritesh-2711/pii-bench.

# 1. Introduction

The detection and redaction of Personally Identifiable Information (PII) is a prerequisite for regulatory compliance under frameworks such as GDPR, HIPAA, and CCPA. Any system processing user-generated or enterprise text must identify spans of text corresponding to sensitive personal data—names, contact details, financial identifiers, credentials—and either redact or otherwise protect them before further processing or storage.

Despite the practical importance of PII detection, the research community lacks a standardized benchmark corpus. Most of the available resources are fragmented: general NER corpora such as CoNLL-2003 cover only four coarse entity types (PER, ORG, LOC, MISC); synthetic PII datasets such as the ai4privacy series define 50+ proprietary label schemas with no cross-dataset alignment; financial corpora such as finer-139 target XBRL disclosure tags that do not map to classical PII categories; and multilingual NER datasets are optimized for entity-level recall rather than PII coverage breadth.

Rule-based systems such as Microsoft Presidio address pattern-matchable PII (e.g., US Social Security Numbers, credit card numbers) but fail on contextually inferred entities—person names in free text, job titles embedded in prose, or usernames present in log-format data.

No published benchmark allows rigorous cross-system comparison across the full taxonomy of PII types that appear in production systems.

We address this gap with PIIBench. Our contributions are:

- A unified corpus of 2,369,883 annotated sequences drawn from ten heterogeneous source datasets, covering general, synthetic, multilingual, and financial domains.
- A normalization pipeline that maps 80+ distinct source label variants to 48 canonical PII entity types under a consistent BIO tagging scheme.
- Stratified 80/10/10 train/validation/test splits with per-source proportional sampling, ensuring that evaluation sets are representative of the full corpus distribution.
- Baseline evaluation of eight published systems—spanning rule-based engines, general-purpose NER transformers, a PII-specific model, and a financial NER specialist—on the held-out test set, establishing a quantitative lower bound for all existing approaches and directly demonstrating the domain-silo problem at scale.

The remainder of this paper is organized as follows.

Section 2 reviews related work.

Section 3 describes the dataset construction methodology.

Section 4 presents corpus statistics.

Section 5 reports baseline benchmark results.

Section 6 discusses limitations and future work.

Section 7 concludes.

# 2. Related Work

## 2.1 Named Entity Recognition Benchmarks

CoNLL-2003 (Sang and De Meulder, 2003) is the canonical NER benchmark, covering four entity types over English and German newswire. OntoNotes 5.0 (Weischedel et al., 2013) extends this to 18 entity types across multiple genres. WikiANN (Pan et al., 2017) provides cross-lingual NER annotations for 282 languages derived from Wikipedia, covering PER, ORG, and LOC. MultiNERD (Tedeschi and Navigli, 2022) extends to 15 fine-grained types. Few-NERD (Ding et al., 2021) introduces 66 fine-grained types for episodic learning.

None of these benchmarks is designed for PII detection; their entity taxonomies include non-sensitive types (ANIMAL, FOOD, VEHICLE) and omit critical PII categories such as financial identifiers, credentials, and government IDs.

## 2.2 PII-Specific Datasets

The ai4privacy dataset series (ai4privacy, 2023) provides large-scale synthetic PII annotations with 54-63 label classes across multiple languages, including a finance-specific FinPII split. Isotonic/pii-masking-200k mirrors this schema under an Apache 2.0 license. The Gretel synthetic finance dataset (Gretel.ai, 2023) covers 29 PII types across 100 financial document formats, validated using LLM-as-judge evaluation. NVIDIA's Nemotron-PII dataset (NVIDIA, 2023) covers 55+ entity types across 50 industries. These datasets each define independent label vocabularies, preventing direct cross-dataset evaluation without normalization.

## 2.3 Financial NER

The FiNER-139 dataset (Loukas et al., 2022) annotates 1.1 million sentences from SEC 10-K/10-Q filings with 139 XBRL financial disclosure tags. While not a classical PII dataset, it provides dense supervision for the FINANCIAL_ENTITY category that is underrepresented in general NER corpora and directly relevant to banking and financial compliance applications.

## 2.4 PII Detection Systems

Microsoft Presidio (Microsoft, 2023) combines regular expression patterns with spaCy-backed NLP to identify 18 PII entity types. It is widely used in production privacy engineering but lacks contextual generalization for entities that do not match pattern templates.

spaCy's en_core_web_lg provides CNN-based NER trained on OntoNotes 5.0.

Transformer-based general NER models on the HuggingFace Hub—dslim/bert-base-NER (CoNLL-2003), Davlan/xlm-roberta-base-wikiann-ner (WikiANN), tomaarsen/span-marker-mbert-base-multinerd (MultiNERD), and tomaarsen/span-marker-bert-base-fewnerd-fine-super (FewNERD)—are widely used but trained exclusively on general NER labels with no PII-specific coverage.

PII-specific models include iiiorg/piranha-v1-detect-personal-information, fine-tuned on ai4privacy-400k, which covers a subset of personal data types but does not generalize across domains. Financial NER is addressed by nbroad/finer-139-xtremedistil-l12-h384, fine-tuned on FiNER-139 XBRL tags, but with no coverage of non-financial PII.

All these systems are siloed to their training domain; none provides comprehensive cross-domain PII coverage. We evaluate all eight systems on PIIBench to quantify this gap directly.

# 3. Dataset Construction

## 3.1 Source Dataset Selection

We select ten publicly available datasets from the HuggingFace Hub based on three criteria: (1) availability of token-level or span-level PII or NER annotations in English; (2) coverage of at least one PII-relevant entity type; and (3) availability under an open license permitting academic use. Table 1 summarizes the selected sources.

| Dataset | Records | Tokens | Avg Tok/Rec | Domain | License |
|---|---|---|---|---|---|
| finer-139 | 1,121,256 | 51,142,079 | 45.6 | Finance (SEC) | CC-BY-SA 4.0 |
| ai4privacy-400k | 325,511 | 16,453,195 | 50.5 | General (synth.) | Custom academic |
| isotonic-pii-200k | 209,261 | 11,147,349 | 53.3 | General (synth.) | Apache 2.0 |
| few-nerd | 188,239 | 4,611,895 | 24.5 | General (Wiki) | CC-BY-SA 4.0 |
| ai4privacy-300k | 177,652 | 23,988,177 | 135.0 | General/Finance | Custom academic |
| multinerd (en) | 131,280 | 2,842,119 | 21.6 | General (Wiki+news) | CC-BY-NC-SA 4.0 |
| nvidia-nemotron | 100,000 | 13,452,035 | 134.5 | General (50+ domains) | CC-BY 4.0 |
| gretelai-finance | 55,940 | 9,435,480 | 168.7 | Finance (banking) | Apache 2.0 |
| wikiann (en) | 40,000 | 321,256 | 8.0 | General (Wikipedia) | CC-BY-SA 3.0 |
| conll2003 | 20,744 | 301,418 | 14.5 | News (Reuters) | Custom non-comm. |
| **Total** | **2,369,883** | **133,695,003** | **56.4** | | |

Table 1: Source datasets included in PIIBench. Records and token counts reflect the splits used in our consolidation pipeline.

The ten sources provide complementary coverage across domains and entity types. Financial corpora (finer-139, gretelai-finance) provide dense annotations for financial entities and structured document formats. Synthetic PII datasets (ai4privacy series, isotonic, nvidia-nemotron) provide broad coverage of personal and credential-type entities. General NER benchmarks (wikiann, multinerd, few-nerd, conll2003) provide contextually realistic text for common entity types such as PERSON, ORG, and LOC.

## 3.2 Annotation Format Normalization

The ten source datasets use heterogeneous annotation formats: token-level BIO tags with integer IDs (few-nerd, conll2003, finer-139), string BIO labels (ai4privacy, isotonic), and character-offset span annotations (gretelai, nvidia-nemotron). We implement three format converters:

- **BIO integer decoder**: maps integer tag IDs to string labels using dataset feature metadata, then normalizes to canonical label strings.
- **Span-to-BIO converter**: tokenizes raw text on whitespace, builds a character-offset to token-index map, and assigns B-/I- labels to tokens covered by each annotated span.
- **XML tag parser**: for the Nemotron-PII dataset, extracts entity spans from tagged text (e.g., <PERSON>John</PERSON>) as a fallback when structured span annotations are absent.

## 3.3 Label Normalization

Source datasets collectively define 80+ distinct label strings for overlapping entity concepts.

We develop a normalization map that collapses these variants into 48 canonical PII entity types. Table 2 illustrates selected normalization rules.

| Canonical Label | Source Variants Normalized |
|---|---|
| **PERSON** | PER, FIRSTNAME, LASTNAME, MIDDLENAME, PREFIX, GENDER, AGE, HEIGHT, EYECOLOR |
| **ORG** | COMPANYNAME, ACCOUNTNAME, COMPANY |
| **LOC** | CITY, STATE, COUNTY, ZIPCODE, NEARBYGPSCOORDINATE, ORDINALDIRECTION |
| **ADDRESS** | STREET, BUILDINGNUMBER, SECONDARYADDRESS |
| **PHONE** | PHONENUMBER, PHONE_NUMBER, PHONEIMEI |
| **CREDIT_CARD** | CREDITCARDNUMBER, CREDITCARDCVV, CREDITCARDISSUER, CREDIT_CARD_NUMBER |
| **ACCOUNT_NUMBER** | ACCOUNTNUMBER, MASKEDNUMBER |
| **CRYPTO_ADDRESS** | BITCOINADDRESS, ETHEREUMADDRESS, LITECOINADDRESS |
| **IP_ADDRESS** | IP, IPV4, IPV6, MAC |
| **JOB** | JOBTITLE, JOBAREA, JOBTYPE |
| **VEHICLE** | VEHICLEVIN, VEHICLEVRM, VEHI |
| **CURRENCY** | CURRENCYCODE, CURRENCYNAME, CURRENCYSYMBOL |
| **FINANCIAL_ENTITY** | XBRL tags (139 variants): Revenue, Assets, LiabilitiesTotal, EPS, ... |

*Table 2: Selected label normalization rules. Source variants from the left are mapped to canonical types on the right. The full normalization map covers 80+ source-specific labels.*

Design decisions in the normalization include:

(1) collapsing demographic sub-types (FIRSTNAME, LASTNAME, MIDDLENAME, PREFIX, GENDER) into PERSON, as these are operationally equivalent for redaction purposes;

(2) retaining FINANCIAL_ENTITY as a distinct category to preserve the specificity of XBRL annotations from finer-139, which would be lost if mapped to ORG; and

(3) mapping all cryptocurrency address variants (BITCOINADDRESS, ETHEREUMADDRESS, LITECOINADDRESS) to CRYPTO_ADDRESS. Labels not covered by the normalization map and matching the camelCase XBRL pattern are mapped to FINANCIAL_ENTITY.

Labels that appear structurally invalid (HTML tags, server configuration keys, version control artifacts) are retained for frequency analysis and subsequently dropped.

## 3.4 Corpus Curation Pipeline

After format normalization, we apply three curation steps:

**Source capping.** The finer-139 source contributes 1,121,256 records—47.3% of the raw corpus—due to the volume of SEC filings. To prevent a single financial-domain source from dominating the training distribution, we apply stratified random sampling to cap finer-139 at 150,000 records during data preparation (the full 1.1M records are retained in the consolidated file; the cap is applied at split time).

**Rare entity suppression.** Entity types with fewer than 500 B-tag mentions in the full corpus are collapsed to the Outside (O) label. This threshold prevents the model from learning associations from near-absent categories that lack sufficient supervision. Of the 80 unique entity types present after normalization, 32 fall below this threshold and are suppressed. These dropped types are primarily markup artifacts (H1, H2, HTML, SCRIPT, UL, OL), system configuration labels (SERVER_CONFIG, WEBSERVERCONFIG, DIRECTORY), and niche biometric identifiers with minimal representation (PATIENTBIOMETRICIDRECORD, VISABIOMETRICDATARECORD).

**Stratified splitting.** An 80/10/10 train/validation/test split is applied independently per source dataset. Records from each source are shuffled with a fixed random seed (42) before splitting, and the resulting splits are interleaved to preserve source diversity within each partition. This procedure ensures that evaluation performance reflects the full distribution of domains and entity types rather than artifacts of any single source.

# 4. Dataset Statistics

## 4.1 Corpus Overview

The consolidated corpus contains 2,369,883 records and 133,695,003 tokens across the ten source datasets. After applying the finer-139 cap of 150,000 records, the effective training corpus spans 1,398,627 records split into train (1,118,899), validation (139,861), and test (139,867) sets. Table 3 summarizes the split statistics.

| Split | Records | % of Total | Notes |
|-------|---------|------------|-------|
| Train | 1,118,899 | 80% | Stratified by source (80% per source) |
| Val | 139,861 | 10% | Stratified by source (10% per source) |
| Test | 139,867 | 10% | Held-out; used for benchmark evaluation |
| Val 1% | 1,398 | 0.1% | Stratified fast-eval subset (1% of val) |
| Test 1% | 1,398 | 0.1% | Stratified fast-eval subset (1% of test) |
| **Total** | **1,398,627** | **100%** | **After finer-139 cap (150k) applied** |

*Table 3: Dataset split statistics. The finer-139 source cap of 150,000 records reduces total records from 2,369,883 to approximately 1,398,627 at training time.*

Average sequence length varies substantially across sources, reflecting their document formats: financial SEC filings (finer-139, gretelai-finance) average 45-168 tokens per record; synthetic PII templates (ai4privacy-300k, nvidia-nemotron) average 134-135 tokens; general NER benchmarks (wikiann, conll2003) average 8-14 tokens per sentence. This diversity in sequence length mirrors the distribution of PII-containing text in production systems, which includes both short, structured records and long unstructured documents.

## 4.2 Entity Type Distribution

The corpus contains 3,352,974 total B-tag entity mentions across 48 canonical types before capping. Table 4 reports the frequency distribution of the top entity types.

| Entity Type | B- Mentions | % of Total | Primary Source(s) |
|---|---|---|---|
| FINANCIAL_ENTITY | 975,734 | 29.1% | finer-139 |
| LOC | 472,846 | 14.1% | wikiann, multinerd, conll2003, few-nerd |
| PERSON | 405,587 | 12.1% | wikiann, multinerd, ai4privacy |
| ORG | 176,102 | 5.3% | wikiann, conll2003, multinerd, few-nerd |
| MISC | 174,929 | 5.2% | conll2003, multinerd, few-nerd |
| DATE | 148,069 | 4.4% | ai4privacy, nvidia-nemotron |
| USERNAME | 117,756 | 3.5% | ai4privacy-400k, isotonic |
| EMAIL | 113,860 | 3.4% | ai4privacy-400k, isotonic |
| TIME | 107,410 | 3.2% | ai4privacy-400k, nvidia-nemotron |
| NAME | 97,100 | 2.9% | nvidia-nemotron, gretelai |
| ADDRESS | 90,433 | 2.7% | ai4privacy, isotonic, gretelai |
| IP_ADDRESS | 87,817 | 2.6% | ai4privacy-400k, isotonic |
| COMPANY | 60,655 | 1.8% | ai4privacy-300k, gretelai |
| STREET_ADDRESS | 41,977 | 1.3% | ai4privacy, gretelai |
| JOB | 38,926 | 1.2% | ai4privacy-400k, isotonic |
| CREDIT_CARD | 33,589 | 1.0% | ai4privacy-400k, isotonic, gretelai |
| ... (32 more types) | — | — | — |
| Total | 3,352,974 | 100% | |

*Table 4: Entity type frequency distribution. Counts reflect the full consolidated corpus before finer-139 capping. The 32 types dropped by frequency filtering are not shown.*

The distribution is heavily skewed: FINANCIAL_ENTITY (from finer-139) accounts for 29.1% of all mentions, followed by LOC (14.1%) and PERSON (12.1%). The tail of the distribution includes 32 entity types with fewer than 500 mentions each, which are suppressed by the rare entity filter. After filtering, 48 entity types are retained, covering a total of 97 BIO label classes (O, plus B- and I- prefixes for each entity type).

The cross-source diversity of entity type coverage is a distinguishing feature of PIIBench. Credential types (USERNAME, PASSWORD, API_KEY) are present only in synthetic PII datasets. Financial identifiers (IBAN, SWIFT_BIC_CODE, CREDIT_CARD) appear primarily in ai4privacy and gretelai-finance. Location entities (LOC, ADDRESS, STREET_ADDRESS) appear across all sources but with different granularities.

## 4.3 Label Normalization Coverage

The normalization map resolves 80+ source-specific label strings to 48 canonical types. The 32 types dropped by frequency filtering are structurally distinct from the retained types: they consist primarily of markup and system configuration labels introduced by the Nemotron-PII dataset's document annotation scheme (e.g., HTML, SCRIPT, SERVER_CONFIG) rather than genuine PII categories. Their suppression does not reduce the coverage of privacy-relevant entity types in the benchmark.

# 5. Benchmark Evaluation

## 5.1 Experimental Setup

We evaluate eight baseline systems on a stratified 1% subset of the test split (1,398 records, 3,383 entity mentions), selected via reservoir sampling to preserve source and entity-type proportions.

Evaluation uses span-level seqeval metrics (Precision, Recall, F1), which require exact match of both the entity span boundaries and the entity type label. We reconstruct each system's predicted character-offset spans back to BIO token labels using a character-to-token alignment procedure, then apply seqeval to the aligned predictions.

The eight baseline systems span three categories.

**Rule-based**: (1) Microsoft Presidio (AnalyzerEngine with default configuration, using spaCy en_core_web_lg as the NLP backend).

**General NER**: (2) spaCy en_core_web_lg (OntoNotes-trained CNN); (3) BERT-base NER [dslim/bert-base-NER], trained on CoNLL-2003; (4) XLM-RoBERTa NER [Davlan/xlm-roberta-base-wikiann-ner], trained on WikiANN English; (5) SpanMarker mBERT [tomaarsen/span-marker-mbert-base-multinerd], trained on MultiNERD; (6) SpanMarker BERT [tomaarsen/span-marker-bert-base-fewnerd-fine-super], trained on FewNERD.

**PII-specific**: (7) Piiranha DeBERTa [iiiorg/piiranha-v1-detect-personal-information], trained on ai4privacy-400k. Financial NER: (8) XtremeDistil FiNER [nbroad/finer-139-xtremedistil-l12-h384], trained on FiNER-139 XBRL tags.

For each system, entity type predictions are mapped to PIIBench canonical labels using a hand-constructed alignment table.

Entity types predicted by any system but not appearing in PIIBench's 48-type taxonomy are discarded. All eight models are publicly available on the HuggingFace Hub at the model IDs listed in Table 5, enabling exact reproduction of the evaluation.

## 5.2 Results

| System | Source Dataset | Type | F1 | Precision | Recall |
|---|---|---|---|---|---|
| Microsoft Presidio | Multiple (rule-based) | Rule-based | 0.1385 | 0.1522 | 0.1271 |
| spaCy en_core_web_lg | OntoNotes 5.0 | General NER | 0.0873 | 0.0636 | 0.1395 |
| SpanMarker mBERT [5] | MultiNERD | General NER | 0.0823 | 0.1477 | 0.0570 |
| SpanMarker BERT [6] | FewNERD | General NER | 0.0877 | 0.1723 | 0.0588 |
| BERT-base NER [1] | CoNLL-2003 | General NER | 0.0657 | 0.0791 | 0.0562 |
| XLM-RoBERTa NER [3] | WikiANN (en) | General NER | 0.0494 | 0.0508 | 0.0482 |
| Piiranha DeBERTa [2] | ai4privacy-400k | PII-specific | 0.0463 | 0.0604 | 0.0375 |
| XtremeDistil FiNER [4] | FiNER-139 | Financial NER | 0.0413 | 0.6990 | 0.0213 |

*Table 5: Baseline system results on the 1,398-record stratified test subset. F1, Precision, and Recall are span-level seqeval metrics (exact span + type match). Source Dataset is the training data the model was originally trained on. Type indicates system category. HuggingFace model IDs for reproducibility: [1] dslim/bert-base-NER; [2] iiiorg/piiranha-v1-detect-personal-information; [3] Davlan/xlm-roberta-base-wikiann-ner; [4] nbroad/finer-139-xtremedistil-l12-h384; [5] tomaarsen/span-marker-mbert-base-multinerd; [6] tomaarsen/span-marker-bert-base-fewnerd-fine-super.*

All eight systems achieve low overall F1 on PIIBench.

Presidio achieves the highest F1 at 0.1385, benefiting from its regex recognizers for structured PII (email, SSN, credit card).

spaCy (0.0873), SpanMarker BERT/mBERT (0.0877 / 0.0823), and BERT-base NER (0.0657) are the next tier, all limited by their CoNLL/OntoNotes/general-NER training distributions.

XLM-RoBERTa NER (0.0494) and Piiranha DeBERTa (0.0463) fall lower: Piiranha, despite being PII-trained, was trained only on ai4privacy-400k and does not generalize to the diverse entity types in PIIBench.

The most striking result is XtremeDistil FiNER (F1=0.0413, Precision=0.6990, Recall=0.0213): its precision is the highest of all systems by a wide margin, but its recall is the lowest. This pattern confirms the domain-silo hypothesis directly. FiNER fires accurately on XBRL financial entities but almost never fires outside its domain, producing near-zero recall on the mixed PIIBench test set. Table 5 reports per-entity results for selected entity types.

| Entity Type | Test Support | Presidio F1 | spaCy lg F1 | Notes |
|---|---|---|---|---|
| FINANCIAL_ENTITY | 873 | 0.0000 | 0.0000 | No system covers XBRL types |
| LOC | 525 | 0.3067 | 0.3070 | Best-performing type; both systems handle well |
| PERSON | 437 | 0.2638 | 0.2644 | Partial coverage; misses synthetic names |
| MISC | 165 | 0.0000 | 0.0000 | Not mapped in either system |
| DATE | 151 | 0.1327 | 0.1020 | High recall but low precision for Presidio |
| ORG | 207 | 0.0000 | 0.0675 | spaCy partially detects; Presidio misses |
| EMAIL | 119 | 0.1277 | 0.0000 | Presidio regex covers; spaCy lacks |
| ADDRESS | 91 | 0.0000 | 0.0000 | Neither system covers structured address |
| COMPANY | 87 | 0.0000 | 0.0000 | Not covered by either baseline |
| IP_ADDRESS | 80 | 0.0000 | 0.0000 | Not covered by either baseline |
| TIME | 126 | 0.0000 | 0.0000 | Not covered by either baseline |
| USERNAME | 109 | 0.0000 | 0.0000 | Not covered by either baseline |
| CREDIT_CARD | 23 | 0.0000 | 0.0000 | Presidio has recognizer but recall = 0 |
| SSN | 16 | 0.0202 | 0.0000 | Presidio pattern partial; US-only |
| IBAN | 13 | 0.0000 | 0.0000 | Neither system covers |
| **... (29 more)** | **—** | **0.0000** | **0.0000** | **All remaining types: F1 = 0 for both** |

Table 6: Per-entity F1 scores for selected entity types. 'Test Support' is the number of ground-truth entity mentions in the 1,398-record test subset. F1 = 0.0000 indicates that the system made no correct predictions for that entity type.

Several patterns are evident from the per-entity breakdown. First, all general NER systems perform reasonably on LOC and PERSON, which are universally covered by CoNLL, OntoNotes, WikiANN, MultiNERD, and FewNERD training data. Second, Presidio's regex recognizers provide coverage of DATE, EMAIL, and PHONE that transformer NER models entirely miss. Third, FINANCIAL_ENTITY—the single most frequent entity type in the test set (873 mentions)— achieves F1 = 0 for all systems except FiNER, which handles only XBRL financial tags and ignores all other entity types. Fourth, credential types (USERNAME, PASSWORD, API_KEY) and financial identifiers (IBAN, CREDIT_CARD, ACCOUNT_NUMBER, SSN) achieve F1 = 0 for all general NER and financial NER systems. Only Piiranha DeBERTa, trained on ai4privacy-400k, produces any predictions for these types, but its coverage remains low due to distribution mismatch with the other nine PIIBench sources.

Taken together, these results confirm the domain-silo problem at a quantitative level: no single existing model covers more than a small fraction of PIIBench entity types effectively. The benchmark presents a substantially harder evaluation challenge than any individual source dataset, and establishes a clear quantitative lower bound that purpose-built, multi-source fine-tuned models must surpass.

# 6. Discussion

## 6.1 Benchmark Validity

A key concern with any benchmark constructed from heterogeneous sources is whether the normalization introduces label noise. We address this in two ways.

First, the normalization map is conservative: we collapse variants only when they refer to operationally equivalent PII concepts (e.g., all person name sub-types map to PERSON).

Second, rare entity types below the 500-mention threshold are suppressed rather than being merged with semantically distant types, preventing spurious label assignments for poorly represented categories. The dropped types are predominantly markup artifacts rather than genuine PII, so their suppression does not degrade benchmark quality.

## 6.2 Domain Coverage and Limitations

PIIBench covers general text (Wikipedia, news), synthetic PII documents, and financial regulatory filings, but does not include medical records, legal contracts, or social media text—domains where PII detection is also critically important. The corpus is predominantly English; while several source datasets include multilingual data, the current benchmark uses English records only. Extending PIIBench to additional domains and languages is a planned direction for future work.

The finer-139 source contributes 47% of raw records before capping. While the 150,000-record cap reduces its dominance, FINANCIAL_ENTITY remains the most frequent entity type in the corpus (29.1% of all mentions).

Researchers using PIIBench for evaluation should be aware that models may be biased toward financial entity detection, and per-entity performance should be analyzed rather than relying solely on micro-average F1.

## 6.3 Reproducibility

The full data construction pipeline—download, consolidation, normalization, and splitting—is implemented in a publicly available codebase. All dataset sources are available on the HuggingFace Hub under open licenses.

The normalization map and random seed (42) are fully specified, enabling exact reproduction of the train/validation/test splits.

## 6.4 Future Work

Several extensions are planned. First, a cross-source generalization experiment—training on N-1 sources and evaluating on the held-out source—will characterize domain transfer within the benchmark and identify which source domains are most challenging for models trained on other domains. Second, we plan to add a multilingual evaluation track using the non-English splits of ai4privacy and gretelai-finance. Third, annotation quality analysis using inter-annotator agreement metrics on a sampled subset of the synthetic data is planned to characterize the noise floor of the benchmark. Fourth, expanding the evaluation suite to include medical and legal NER models—domains currently absent from PIIBench sources—is a planned direction for broadening coverage.

# 7. Conclusion

We presented PIIBench, a unified benchmark corpus for PII detection comprising 2,369,883 annotated sequences across 48 canonical entity types, drawn from ten heterogeneous public datasets.

We described a reproducible normalization pipeline that resolves 80+ incompatible label schemes into a standardized BIO format, and stratified split procedures that preserve source-level diversity in evaluation sets.

Baseline evaluation across eight published systems—spanning rule-based engines, general NER transformers, a PII-specific model, and a financial NER specialist—demonstrates that all existing models achieve F1 below 0.14 on the benchmark.

The results directly quantify the domain-silo problem: **no single model covers more than a small fraction of PIIBench entity types**, and the highest-precision system (XtremeDistil FiNER, P=0.699) achieves near-zero recall (R=0.021) outside its financial domain. PIIBench provides the research community with a rigorous, multi-domain evaluation testbed for PII detection systems and a large-scale training corpus for fine-tuning token classification models.